\documentclass[letterpaper, 10pt, conference]{ieeeconf}  % Comment this line out if you need a4paper
\IEEEoverridecommandlockouts                              % This command is only needed if you want to use the \thanks command
\usepackage{graphics} % for pdf, bitmapped graphics files
\usepackage{epsfig} % for postscript graphics files
\usepackage{times} % assumes new font selection scheme installed
\usepackage{amsmath} % assumes amsmath package installed
\usepackage{amssymb}  % assumes amsmath package installed
\usepackage[T1]{fontenc}  % handle Command DH error
\usepackage[misc]{ifsym}  % add e-mail icon to corresponding author
\usepackage[noadjust]{cite}
\usepackage{multirow, multicol}
\usepackage{array}
\usepackage{hyperref}
\hypersetup{hidelinks}

\title{\LARGE \bf GraNet: A Multi-Level Graph Network for 6-DoF Grasp Pose Generation in Cluttered Scenes }

\author{Haowen Wang, Wanhao Niu, Chungang Zhuang$^{\dag}$ % <-this % stops a space
\thanks{All authors are with the Institute of Robotics, School of Mechanical Engineering, Shanghai Jiao Tong University, Shanghai, China. Email: \texttt{\{\href{mailto:wang.h.w@sjtu.edu.cn}{wang.h.w}\,,\,\href{mailto:nwh1093412390@sjtu.edu.cn}{nwh1093412390}\,,\,\href{mailto:cgzhuang@sjtu.edu.cn}{cgzhuang}\}@sjtu.edu.cn}.} %
\thanks{$^{\dag}$\,Corresponding author: Chungang Zhuang.}%
}

\begin{document}

\maketitle
\thispagestyle{empty}
\pagestyle{empty}

%%%%%%%%%%%%%%%%%%%%%%%%%%%%%%%%%%%%%%%%%%%%%%%%%%%%%%%%%%%%%%%%%%%%%%%%%%%%%%%%
\begin{abstract}

% Importance
6-DoF object-agnostic grasping in unstructured environments is a critical yet challenging task in robotics. 
% Previous works
Most current works use non-optimized approaches to sample grasp locations and learn spatial features without concerning the grasping task.
% Overview of GraNet
This paper proposes GraNet, a graph-based grasp pose generation framework that translates a point cloud scene into multi-level graphs and propagates features through graph neural networks.
% Point selection & multi-graph
By building graphs at the \textit{scene level}, \textit{object level}, and \textit{grasp point level}, GraNet enhances feature embedding at multiple scales while progressively converging to the ideal grasping locations by learning.
% Benefits
Our pipeline can thus characterize the spatial distribution of grasps in cluttered scenes, leading to a higher rate of effective grasping.
Furthermore, we enhance the representation ability of scalable graph networks by a structure-aware attention mechanism to exploit local relations in graphs.
% Experiments: GraspNet-1Billion
Our method achieves state-of-the-art performance on the large-scale GraspNet-1Billion benchmark, especially in grasping unseen objects (+11.62 AP).
% Experiments: Real
The real robot experiment shows a high success rate in grasping scattered objects, verifying the effectiveness of the proposed approach in unstructured environments.

\end{abstract}

%%%%%%%%%%%%%%%%%%%%%%%%%%%%%%%%%%%%%%%%%%%%%%%%%%%%%%%%%%%%%%%%%%%%%%%%%%%%%%%%
\section{INTRODUCTION}

% Importance to challenges
Robotic grasping is a widely researched field in robotics for its pivotal role in automated assembly. However, when robots are confronted with increasingly complex scenarios, most grasp pose generation methods fail to demonstrate sufficient adaptability to the environment, resulting in conflicted or unreliable grasps.
% Traditional methods: model-based or 3-DoF
Traditional grasp pose generation methods \cite{zeng_multi-view_2017, wang_densefusion_2019} focus on the localization of known models by the pose estimation network, ignoring the design of grasp configurations. These methods cannot be generalized to scenes with unknown objects. Some 3-DoF model-free methods \cite{pinto_supersizing_2016, kumra_robotic_2017} are only applicable to scenes where the end-effector grasps vertically downward and thus cannot determine the best direction to approach the target.

% Current: point cloud and 6-DoF grasping
Point clouds provide a fine-grained description of the 3D space and have been widely used in vision tasks. Using point clouds, 6-DoF grasp pose generation methods \cite{mousavian_6-dof_2019, murali_6-dof_2020, lou_learning_2020} allow for a better understanding of spatial structures and are therefore able to generate conflict-free grasp poses from multiple directions. Most studies \cite{liang_pointnetgpd_2019, ni_pointnet_2020, qin_s4g_2020, wang_graspness_2021} use PointNet++ \cite{qi_pointnet_2017-1} as the backbone to extract multi-scale features in a hierarchical manner.
% PointNet++ problem
PointNet++ uses sampling layers to aggregate information from neighbor points but ignores the feature relations between points, which may have an impact on tasks such as grasp prediction that are sensitive to relative relationships. Moreover, the "down-sampling + up-sampling" learning approach makes it difficult to propagate shallow features to distant points.

\begin{figure}[t]
    \centering
    \includegraphics[width=\linewidth]{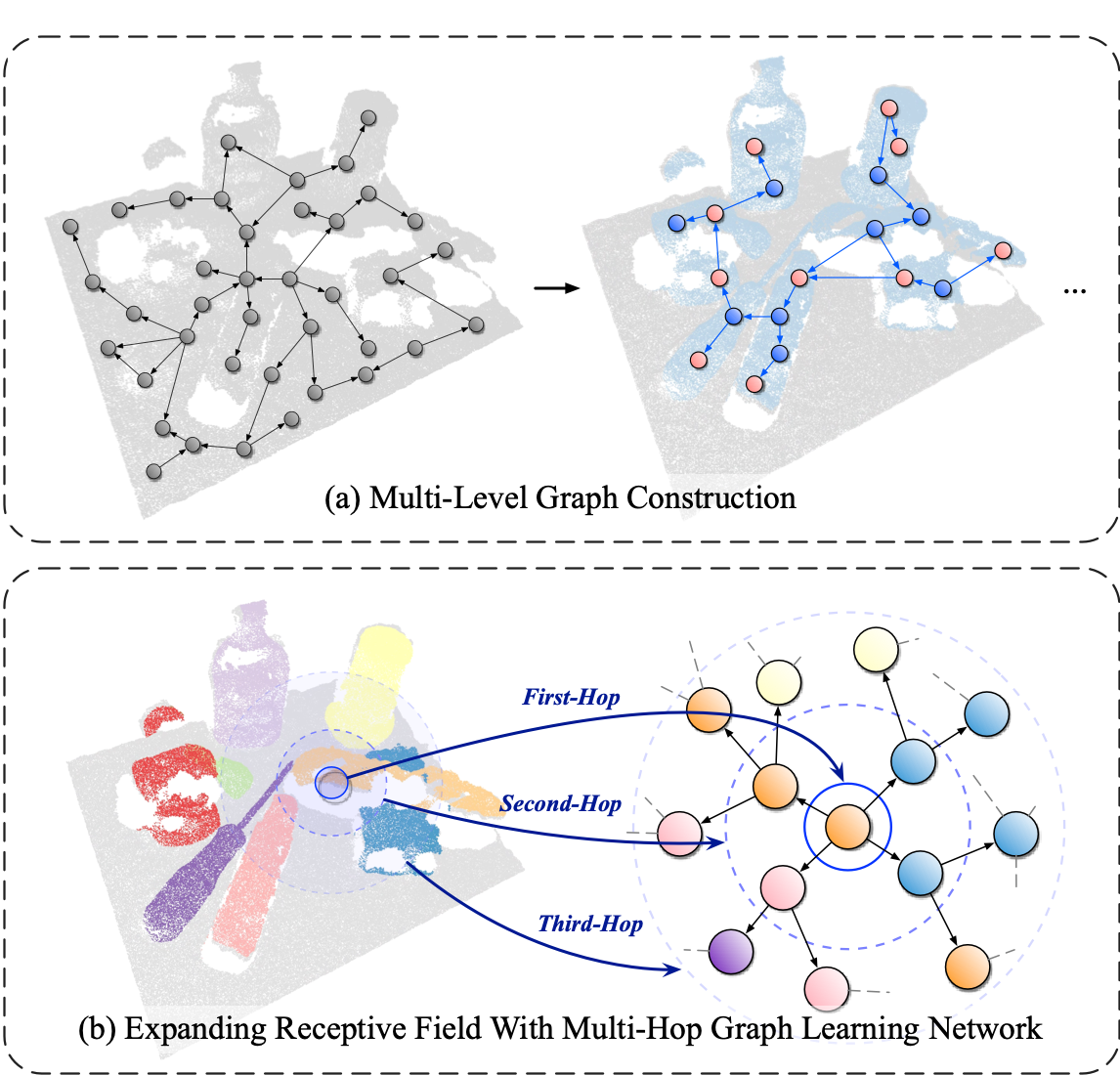}
    \caption{An illustration of our multi-level and multi-hop graph learning network. (a) We construct graphs at different depths of the network, creating a cascading learning effect to locate ideal grasp locations. (b) Our graph network continuously expands the perceptive field during propagation, which is different from previous ways of obtaining features by sampling.}
    \label{fig:toy_illustration}
    \vspace{-1.5em}
\end{figure}

% Current: sample and evaluate
GraspNet-1Billion \cite{fang_graspnet-1billion_2020} provides a large dataset containing rich grasp annotations and designs an end-to-end network for regression-based grasp pose prediction. This type of network first samples grasp points in the scene and then evaluates whether the grasp is feasible at each location.
% Sampling problem
However, uniform sampling leads to the fact that not all sampled points are optimal, \textit{e.g.} background points cannot be used as grasp centers. This non-learning sampling approach introduces redundant computations at non-grasping locations while reducing the proportion of feasible grasps.

% My network
To address the above problems, this work builds graphs based on the point cloud, designing graph networks to learn multi-scale spatial features.
% Multi-level graph and learning-based point selection
Instead of applying a separate backbone network to generate generic features and then sampling them uniformly, our framework performs strategic point selection during feature extraction in conjunction with the grasping task. Whenever we sample nodes in the original graph, we select those points that are ideally located on the object. After each point selection process, we construct new graphs to reflect the feature associations in the current point set, forming a multi-level graph network.
As shown in Fig.\ref{fig:toy_illustration}(a), for example, the original graph (left) covers the entire point cloud scene, and after network learning and point selection, the new graph (right) will contain only the surface points on the target. The new graph contains more spatial and semantic information in each node, for it is in the posterior position of the network.
% Graph network
We utilize graph networks to capture spatial features embedded in the point cloud and achieve the expansion of the perceptual field through the information transmission between nodes, as shown in Fig.\ref{fig:toy_illustration}(b). We consider the multi-hop connectivity around each node and use it to express the local features at that point. Since edges can reflect the relative relations between nodes, our graph network can learn similarities and differences more efficiently.

% Contributions
In summary, our main contributions are as follows:
\begin{itemize}
    \item We propose an end-to-end \textbf{Gra}ph-based \textbf{Gra}sp pose generation network dubbed \textbf{Gra}Net that adopts a learning strategy to choose grasp points and deeply integrates the grasping task into the feature extraction network. Our framework can effectively generate grasp poses in scattered scenes containing unseen objects, achieving state-of-the-art performance on the GraspNet-1Billion dataset.
    \item We design a multi-hop graph feature embedding network with channel attention to enhance the feature learning of graph networks in local regions.
\end{itemize}

%%%%%%%%%%%%%%%%%%%%%%%%%%%%%%%%%%%%%%%%%%%%%%%%%%%%%%%%%%%%%%%%%%%%%%%%%%%%%%%%
\section{RELATED WROK}

\subsection{6-DoF Grasp Pose Generation}

% Model-based & 3-DoF -> Model-free & 6-DoF
Model-based grasp pose generation methods rely on the prior knowledge of targets and usually transfer existing grasps to the scene after applying object pose estimation \cite{wong_segicp_2017, tekin_real-time_2018, dong_ppr-netpoint-wise_2019}, which is unable to grasp novel objects and explore potential grasp domain in challenging environments. Many recent methods focus on model-free 6-DoF grasp detection using depth image or point cloud as input \cite{ni_pointnet_2020, gou_rgb_2021, wen_catgrasp_2022}. Compared with rectangle-based \cite{jiang_efficient_2011, redmon_real-time_2015} or pixel-based \cite{wang_efficient_2019, kumra_antipodal_2020} 3-DoF grasping, 6-DoF grasping can approach the target from different directions in space, which can better deal with scattered and stacked grasp scenarios.

% Model-free networks review
Object-agnostic approaches usually sample candidate grasp poses from the scene and evaluate grasp quality according to different metrics. GPD \cite{pas_grasp_2017} generates grasp candidates using hand-crafted surface features from the normalized point cloud and constructs a CNN-based classifier to evaluate grasp quality. PointNetGPD \cite{liang_pointnetgpd_2019} extends this work by using PointNet \cite{qi_pointnet_2017} to capture local geometric features inside the gripper. 6-DoF GraspNet \cite{mousavian_6-dof_2019} designes grasp sampling rules via variational auto-encoder, introducing PointNet++ \cite{qi_pointnet_2017-1} to evaluate grasp qualities with a posterior optimization module. $\rm S^4G$ \cite{qin_s4g_2020} also uses PointNet++ as the backbone network to directly generate 6-DoF grasp poses and confidence scores at key points. Contact-GraspNet \cite{sundermeyer_contact-graspnet_2021} maps the distribution of 6-DoF grasps to their corresponding contact points \cite{alliegro_end--end_2022}, decomposing the grasp generation problem into contact point classification and grasping direction estimation. REGNet \cite{zhao_regnet_2021} builds a three-stage network to regress grasp poses at key point locations and learns the local region features of key points to refine the predicted poses.

% GraspNet-1Billion series
GraspNet-1Billion \cite{fang_graspnet-1billion_2020} constructs a large-scale dataset containing over one billion grasp poses and proposes a benchmark network to obtain grasp position and approach direction in a decoupled manner. \cite{lu_hybrid_2022} proposes a hybrid physical metric to refine the grasp confidence scores in GraspNet-1Billion. GVN \cite{tian_vote_2021} introduces voting mechanism to fine-tune the grasping center location. \cite{li_simultaneous_2021} combines target semantic information to generate point-wise grasp pose predictions and conflict judgments. TransGrasp \cite{liu_transgrasp_2022} uses Transformer to resolve the feature associations of distant points in the scene to enhance the perception of global characteristics. 

\subsection{Graph Neural Networks in Robotics}

% GNNs in CV
Graph neural networks (GNN) have shown promising performance in processing non-Euclidean structured data \cite{wu_comprehensive_2021} and have been applied to multiple fields such as natural language processing \cite{yao_graph_2019} and social network modeling \cite{bian_rumor_2020}. In robotics, more and more studies are constructing graph representations among physical data to model the intrinsic association between signals \cite{garcia-garcia_tactilegcn_2019, funabashi_multi-fingered_2022}.
% Graph networks in grasping
\cite{iriondo_affordance-based_2021} utilizes deep GNNs as the backbone network to generate point cloud features and predict the grasping affordance of each sampled point. \cite{lou_learning_2022} extracts object-level geometric features by a 3D convolutional autoencoder and constructs grasp graphs to represent the spatial relationship between objects and grasp poses. The above studies used GNNs in one part of the process but did not integrate graphs with the grasping task. Our paper builds graphs at multiple levels and uses GNNs to capture multiple relationships.

%%%%%%%%%%%%%%%%%%%%%%%%%%%%%%%%%%%%%%%%%%%%%%%%%%%%%%%%%%%%%%%%%%%%%%%%%%%%%%%%
\section{PROPOSED APPROACH}

%%%%%%%%%%%%%%%%%%%%%%%%%%%%%%%%%%%%%%%
\subsection{Problem Statement}

Given a partial point cloud $\mathcal{P}=\{\boldsymbol{p}_i\in\mathbb{R}^3\}_{i=1}^N$ of a scene, we aim to learn the mapping $\mathcal{M}$ from point cloud $\mathcal{P}$ to grasp space $\mathcal{G}=\{\boldsymbol{g}_i\in\mathbb{R}^6\}_{i=1}^M$. Each feasible grasp can be represented by $\boldsymbol{g}=\{\boldsymbol{R},\boldsymbol{T}\}\in SE(3)$ where $\boldsymbol{R}\in SO(3)$ and $\boldsymbol{T}\in\mathbb{R}^3$ are the rotation and translation of the grasp. We learn the complex mapping $\mathcal{M}:\mathcal{P}\rightarrow\mathcal{G}$ by the proposed multi-level graph network.

%%%%%%%%%%%%%%%%%%%%%%%%%%%%%%%%%%%%%%%
\begin{figure*}[t]
    \centering
    \includegraphics[width=\linewidth]{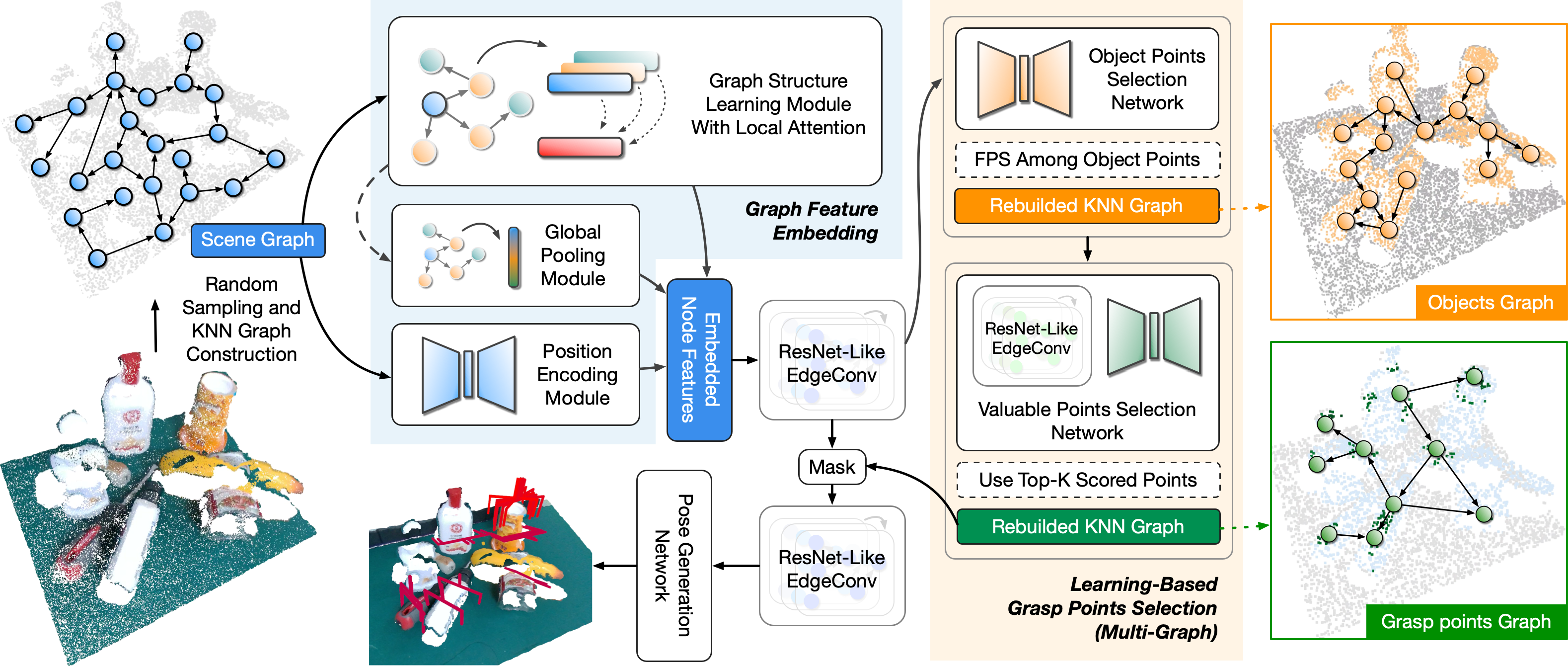}
    \vspace{-1.5em}
    \caption{The architecture of GraNet. Graph feature embedding network takes scene graph as input, incorporating attention mechanisms to fuse multi-scale information from graph structure. Learning-based grasp point selection network build graphs at different feature levels to achieve a progressive focus effect on valuable grasp points. Our pipeline propagates node features across multiple graphs to predict grasp pose configurations in an end-to-end manner. }
    \label{fig:pipeline}
    \vspace{-1em}
\end{figure*}

\subsection{Pipeline Overview}

The overall architecture of GraNet is illustrated in Fig. \ref{fig:pipeline}. GraNet consists of three subnetworks: graph feature embedding network (GFE), learning-based grasp point selection network (GPS) and grasp pose generation network (GPG).

GFE outputs embedded features from three encoding modules to represent graph relations at different scales. These features are then passed into GPS, producing masks on the origin scene to locate grasps on valuable positions. Multi-level graphs are constructed in this process. Finally, GPG generates grasp configuration parameters at selected locations. We construct ResNet-like graph networks between submodules based on EdgeConv \cite{wang_dynamic_2019}. More details about GraNet will be described below.

%%%%%%%%%%%%%%%%%%%%%%%%%%%%%%%%%%%%%%%
\subsection{Graph Feature Embedding Network (GFE)}

\subsubsection{Graph Structure Learning With Local Attention}

% SIGN and local attention introducing
To extract features from more graph nodes at the early stage of the network, we design a feature embedding network based on "flattened" SIGN \cite{frasca_sign_2020}, which applies graph convolutions in parallel to aggregate multi-hop features. SIGN is faster and computationally less expensive than traditional "stacking" graph networks but does not consider the importance of multi-hop features in local regions. We enhance the feature learning capability of SIGN by introducing a local attention mechanism.

% Formula of SIGN
Let $\mathcal{G}=\{\mathcal{V}, \mathcal{E}\}$ denote a directed graph with node set $\mathcal{V}$ and edge set $\mathcal{E}$. Let $\mathbf{W}\in\mathbb{R}^{N\times N}$ be the adjacency matrix where $w_{ij}>0$ if $(\mathcal{V}_i,\mathcal{V}_j)\in\mathcal{E}$. Let $\mathbf{D}\in\mathbb{R}^{N\times N}$ be the diagonal degree matrix where $d_{ii}$ is the degree of node $\mathcal{V}_i$. The symmetrically normalized adjacency matrix $\mathbf{A}\in\mathbb{R}^{N\times N}$ of hop $p$ can be calculated as
\begin{equation}
    \mathbf{A_p} = \mathbf{A}^p\ ,\ \mathbf{A} = \mathbf{D}^{-\frac{1}{2}}\mathbf{W}\mathbf{D}^{-\frac{1}{2}}
\end{equation}
and the learned feature of hop $p$ is
\begin{equation}
    \mathbf{H_p} = \varphi_p \left(\mathbf{A_p}\mathbf{X}\right)
\end{equation}
where $\varphi_p:\mathbb{R}^{d}\rightarrow\mathbb{R}^{f}$ being a learnable function and $\mathbf{X}\in\mathbb{R}^{N\times d}$ is the input node feature of dimension $d$.

\begin{figure}[t]
    \centering
    \includegraphics[width=\linewidth]{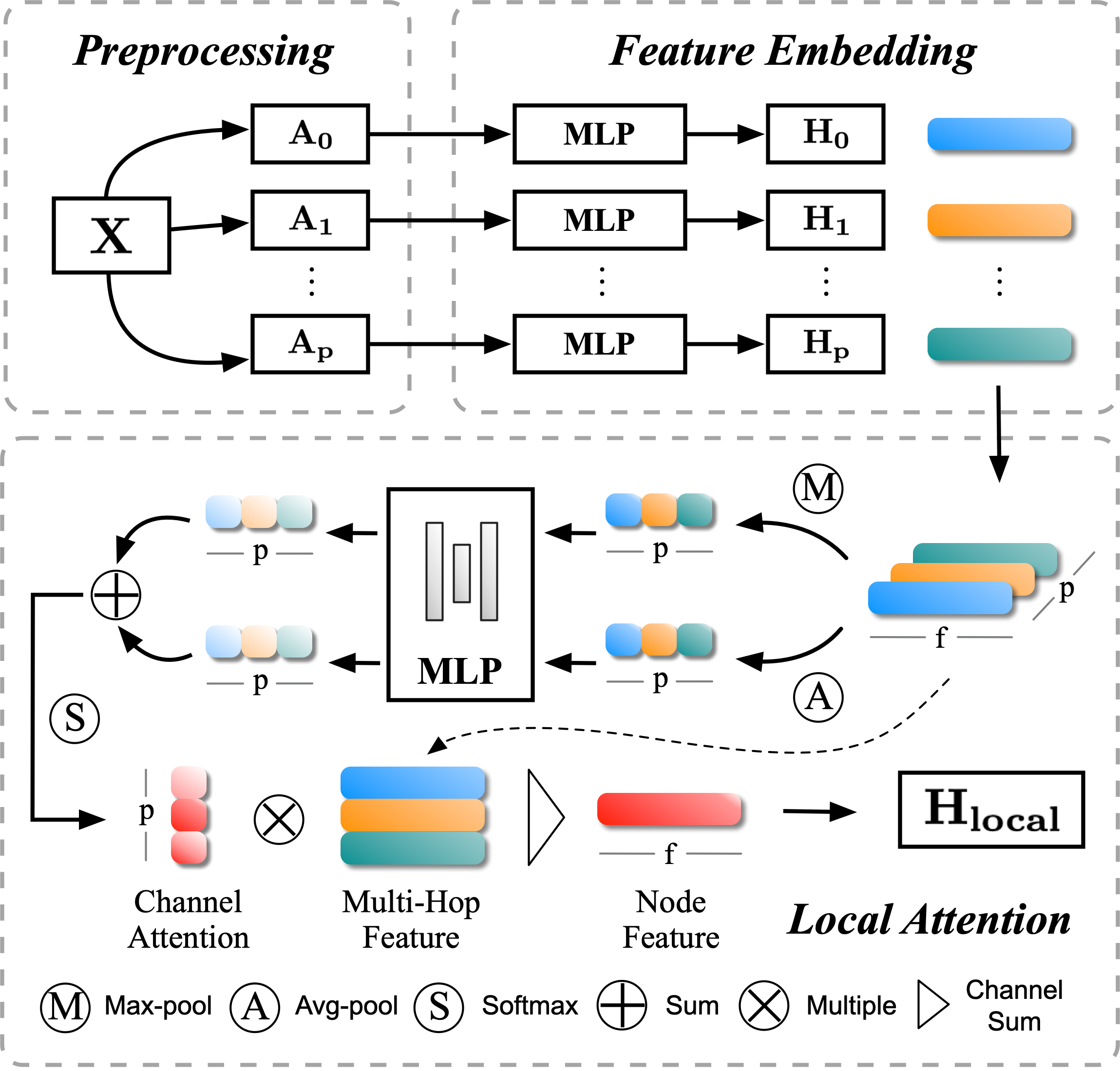}
    \vspace{-1.5em}
    \caption{Graph structure learning module based on SIGN \cite{frasca_sign_2020}. We design a structure-aware attention mechanism to strengthen the local features around each graph node.}
    \label{fig:sign}
    \vspace{-1.5em}
\end{figure}

% Channel attention
After obtaining multi-hop features, we use hop attention shown in Fig. \ref{fig:sign} to generate local features:
\begin{equation}
    \mathbf{S} = \mathop{\mathrm{softmax}}\limits_p\left(\psi_m(\mathop{\mathrm{max}}\limits_f\left(\mathbf{H_{cat}}\right)) + \psi_a(\mathop{\mathrm{avg}}\limits_f\left(\mathbf{H_{cat}}\right))\right)
\end{equation}
\begin{equation}
    \mathbf{H_{local}} = \widehat{\mathbf{S}} \odot \mathbf{H_{cat}}
    \label{equation:h_local}
\end{equation}
where $\mathbf{H_{cat}}\in\mathbb{R}^{N\times fp}$ is the concatenation of multi-hop features, $\mathrm{max}$ and $\mathrm{avg}$ are pooling layers across feature dimension, $\widehat{\mathbf{S}}\in\mathbb{R}^{N\times fp}$ expands $\mathbf{S}$ on dimension $f$, $\odot$ is the element-wise product, $\psi:\mathbb{R}^{p}\rightarrow\mathbb{R}^{p}$ are learning functions on pooled features, we set $\psi_m$ same with $\psi_a$ in practice. The obtained attentional features $\mathbf{H_{local}}\in\mathbb{R}^{N\times f}$ express the influence of nearby multi-hop information with stronger local geometry learning ability. We hypothesize that local features are crucial for the grasp pose detection problem, which will be elucidated in the subsequent sections.

\subsubsection{Global Pooling And Position Encoding}

The scalable graph structure learning module with attention mechanism describes the local features but loses the conceptual understanding of the scene. Classical point cloud processing networks \cite{qi_pointnet_2017} usually introduce global features to encode the scene as a whole. The fusion of local and global knowledge can extend the gradient update to a broader range, balancing diversity and uniformity. To make the global pooling conform to the graph structure, we adopt a simple method in \cite{li_gated_2017} to encode the global feature in an attention-based manner:
\begin{equation}
    \boldsymbol{h}_{global} = \sum_{n=1}^{N}{\left(\mathop{\mathrm{softmax}}\limits_{N}\left(\theta_{g}\left(\mathbf{H_{local}}\right)\right) \odot \theta_{f}\left(\mathbf{H_{local}}\right)\right)}
    \label{equation:h_global}
\end{equation}
where $\theta:\mathbb{R}^f\rightarrow\mathbb{R}^{1}$ are MLPs. For simplicity, we set $\theta_{f}$ same with $\theta_{g}$.  Furthermore, we design the position encoding module to emphasize the impact of position relations on grasping:
\begin{equation}
    \mathbf{H_{pos}} = \sigma\left(\mathbf{X}\right)
    \label{equation:h_pos}
\end{equation}
where $\sigma:\mathbb{R}^{3}\rightarrow\mathbb{R}^{f}$.

The final node features $\mathbf{H}=\left\{\boldsymbol{h}_i\in\mathbb{R}^{f\times 3}\right\}_{i=1}^{N}$ obtained by the graph feature embedding network is the concatenation of equation (\ref{equation:h_local}), (\ref{equation:h_global}) and (\ref{equation:h_pos}). The global feature $\boldsymbol{h}_{global}$ is appended to each node.

%%%%%%%%%%%%%%%%%%%%%%%%%%%%%%%%%%%%%%%
\subsection{Learning-Based Grasp Points Selection Network (GPS)}

The learning-based grasp points selection strategy selects areas on the target surface that are suitable for grasping. GPS can be further divided into object points selection network (OPS) and valuable points selection network (VPS). OPS filters out nodes located on the surface of objects, as background points do not yield effective grasps. VPS analyzes the spatial distribution of grasps and decides which areas are suitable for implementing grasping. After each filtering, we build graphs based on the new grasp point candidates, forming a multi-graph architecture. GPS does not rely on prior semantic information of objects and only infers the reliability of grasping based on spatial features, which can be applied to weakly textured objects.

\subsubsection{Object Points Selection Network (OPS)}

The features obtained after GFE already contain rich local information. We believe that for binary classification problems such as selecting front attractions, local knowledge within the appropriate scale is critical for modeling nonlinear relations. Using the features learned by GFE as input, OPS trains a simple binary classifier $\zeta_{obj}$ to discriminate whether each node belongs to the object surface. We sample in the surface point set and recreate a new graph $\mathcal{G}_{obj} = \{\mathcal{V}_{obj}, \mathcal{E}_{obj}\}$ where
\begin{equation}
    \mathcal{V}_{obj} = \textrm{FPS}\left(\mathrm{argmax}\left(\zeta_{obj}(\mathbf{H})\right)\overset{mask}{\longrightarrow}\mathcal{V}\right)
\end{equation}

\subsubsection{Valuable Points Selection Network (VPS)}

The optimal location for object grasping is closely related to the scene condition, e.g. heavily occluded areas may lead to physical conflicts and are therefore not suitable for stable grasping. We calculate the average scores of all grasps at each sampled object node and divide the scores into $M$ levels as the \textit{degree of value (DoV)} at that place:
\begin{equation}
    \hat{s}_i = \frac{\sum_{g\in\boldsymbol{g}_i}{s(g)}}{\mid \boldsymbol{g}_i \mid} \ ,\ i=1,...,N_{obj}
\end{equation}
\begin{equation}
    DoV_{i} = \left\lfloor \frac{\hat{s}_i - \mathrm{min}(\mathcal{S})}{\mathrm{max}(\mathcal{S}) - \mathrm{min}(\mathcal{S})} \times M \right\rfloor  \ ,\ i=1,...,N_{obj}
\end{equation}
where $s\left(\cdot\right)$ is the score function of grasps, $\boldsymbol{g}_i$ represents all grasps annotated for node $\mathcal{V}_i$, $\mathcal{S}$ is the collection of $s_i$ in one scene.

VPS classifies the value of each point using a ResNet-like graph network $\mathcal{F}$ and a multi-classifier $\zeta_{val}$. Compared with regression, classification is more conducive to learning.
% Grasping in real scenes is usually area-oriented due to the size of the end-effector. To characterize the grasping value in regions, we use the average $DoV$ in the spherical domain centered at each point as the final score. 
We select top-\textit{k} scored points as final grasp points and build graph $\mathcal{G}_{val} = \{\mathcal{V}_{val}, \mathcal{E}_{val}\}$ among them, where
\begin{equation}
    \mathcal{V}_{val} = \textrm{Top-}k\left(\mathrm{argmax}\left(\zeta_{val}(\mathcal{F}(\mathbf{H}))\right)\overset{score}{\longrightarrow}\mathcal{V}_{obj}\right)
\end{equation}

As shown in Fig. \ref{fig:pipeline}, in the VPS and around the GPS, we use ResNet-like EdgeConv for graph feature learning, which can be represented as:
\begin{equation}
    \boldsymbol{h}_{i}^{(l+1)_{res}} = \mathop{\mathrm{max}}\limits_{\left(\mathcal{V}_i,\mathcal{V}_j\right)\in\mathcal{E}} \left(\omega_r\left(\boldsymbol{h}_j^{(l)} - \boldsymbol{h}_i^{(l)}\right) + \omega_c\boldsymbol{h}_i^{(l)}\right)
\end{equation}
\begin{equation}
    \mathcal{G}^{(l+1)}_{*} = \mathcal{G}^{(l+1)_{res}}_{*} + \mathcal{G}^{(l)}_{*}
\end{equation}
where $\boldsymbol{h}_i^{(l)}$ is the feature of $\mathcal{V}_i$ in $\mathcal{G}^{(l)}_{*}$, $*$ indicates $obj$ or $val$, $\omega$ are learnable functions.

%%%%%%%%%%%%%%%%%%%%%%%%%%%%%%%%%%%%%%%
\subsection{Grasp Pose Generation Network (GPG)}

After obtaining $\mathcal{G}_{val}$ and corresponding node features through GPS, we adopt a grasp parameter generation method similar to  \cite{fang_graspnet-1billion_2020}, which decouples the orientation $\boldsymbol{R}$ into approaching direction and in-plane rotation. GPG classifies approaching vectors into $V$ predefined viewpoints and selects the highest-scored direction to approach. In-plane rotation, width and depth, which are parameters closely related to local geometric features, are generated directly from low-level features with the help of cylinder query. 
% Classification: approach vector
Different from \cite{fang_graspnet-1billion_2020}, we will not perform foreground segmentation again in this stage because we have already selected object points in the previous GPS network.

%%%%%%%%%%%%%%%%%%%%%%%%%%%%%%%%%%%%%%%

\renewcommand\arraystretch{1.1}
\begin{table*}[t]
    \centering
    \caption{Evaluation for different models based on GraspNet-1Billion benchmark captured by Realsense/Kinect respectively.}
    \label{tab:result}
    \resizebox{\textwidth}{!}{
    \begin{tabular}{c|c|c|c|c|c|c|c|c|c}
        \hline
        \multirow{2}{*}{Methods} & \multicolumn{3}{c|}{Seen} & \multicolumn{3}{c|}{Unseen} & \multicolumn{3}{c}{Novel} \\
        \cline{2-10}
         & $\mathrm{AP}$ & $\mathrm{AP_{0.8}}$ & $\mathrm{AP_{0.4}}$ & $\mathrm{AP}$ & $\mathrm{AP_{0.8}}$ & $\mathrm{AP_{0.4}}$ & $\mathrm{AP}$ & $\mathrm{AP_{0.8}}$ & $\mathrm{AP_{0.4}}$ \\
        \hline
        GG-CNN \cite{morrison_closing_2018} & 15.48/16.89 & 21.84/22.47 & 10.25/11.23 & 13.26/15.05 & 18.37/19.76 & 4.62/6.19 & 5.52/7.38 & 5.93/8.78 & 1.86/1.32 \\
        Chu \textit{et al.} \cite{chu_real-world_2018} & 15.97/17.59 & 23.66/24.67 & 10.80/12.74 & 15.41/17.36 & 20.21/21.64 & 7.06/8.86 & 7.64/8.04 & 8.69/9.34 & 2.52/1.76 \\
        GPD \cite{pas_grasp_2017} & 22.87/24.38 & 28.53/30.16 & 12.84/13.46 & 21.33/23.18 & 27.83/28.64 & 9.64/11.32 & 8.24/9.58 & 8.89/10.14 & 2.67/3.16 \\
        PointNetGPD \cite{liang_pointnetgpd_2019} & 25.96/27.59 & 33.01/34.21 & 15.37/17.83 & 22.68/24.38 & 29.15/30.84 & 10.76/12.83 & 9.23/10.66 & 9.89/11.24 & 2.74/3.21 \\
        GraspNet-baseline \cite{fang_graspnet-1billion_2020} & 27.56/29.88 & 33.43/36.19 & 16.95/19.31 & 26.11/27.84 & 34.18/33.19 & 14.23/16.62 & 10.55/11.51 & 11.25/12.92 & 3.98/3.56 \\
        RGBMatter \cite{gou_rgb_2021} & 27.98/32.08 & 33.47/39.46 & 17.75/20.85 & 27.23/30.40 & 36.34/37.87 & 15.60/18.72 & 12.25/\textbf{13.08} & 12.45/13.79 & 5.62/\textbf{6.01} \\
        Li \textit{et al.} \cite{li_simultaneous_2021} & 36.55/\;\;\;\,-\;\;\;\, & 47.22/\;\;\;\,-\;\;\;\, & 19.24/\;\;\;\,-\;\;\;\, & 28.36/\;\;\;\,-\;\;\;\, & 36.11/\;\;\;\,-\;\;\;\, & 10.85/\;\;\;\,-\;\;\;\, & 14.01/\;\;\;\,-\;\;\;\, & 16.56/\;\;\;\,-\;\;\;\, & 4.82/\;\;\,-\;\;\, \\
        TransGrasp \cite{liu_transgrasp_2022} & 39.81/35.97 & 47.54/41.69 & \textbf{36.42}/31.86 & 29.32/29.71 & 34.80/35.67 & 25.19/24.19 & 13.83/11.41 & 17.11/\textbf{14.42} & 7.67/5.84 \\
        \hline
        Ours (w/o GPS) & 33.05/31.97 & 40.12/38.52 & 24.73/22.68 & 29.40/28.15 & 35.87/33.20 & 21.33/20.86 & 11.94/10.25 & 15.39/13.28 & 7.05/4.96 \\
        Ours & \textbf{43.33}/\textbf{41.48} & \textbf{52.56}/\textbf{49.84} & 34.03/\textbf{33.86} & \textbf{39.98}/\textbf{35.29} & \textbf{48.66}/\textbf{43.15} & \textbf{32.00}/\textbf{26.89} & \textbf{14.90}/11.57 & \textbf{18.66}/14.31 & \textbf{7.76}/5.24 \\
        \hline
    \end{tabular}}
\end{table*}

\subsection{Overall Loss}

For the two-stage classification task in GPS, we aim to learn the correct masks of the scene by:
\begin{equation}
    L_{GPS} = \frac{1}{N}\sum_{i=1}^{N}{L_{cls}\left(\hat{o}_i, o_i\right)} + \frac{1}{N_{obj}}\sum_{i=1}^{N_{obj}}{L_{cls}\left(\hat{v}_i, v_i\right)}
\end{equation}
where $o_i$ is the ground-truth binary mask of $\mathcal{V}$ where $o_i=1$ if the point belong to object surface, $v_i$ is the $DoV$ category for nodes in $\mathcal{V}_{obj}$, $\hat{o}_i$ and $\hat{v}_i$ are corresponding predicted values, $L_{cls}$ denote cross entropy loss function.

Together with GPG, the overall loss is formulated as:
\begin{equation}
    L = \lambda_{1}L_{GPS} + \lambda_{2}L_v + \lambda_{3}\left(L_s + L_r + L_w\right)
\end{equation}
where $L_v$ and $L_r$ are the classification loss for approaching directions and in-plane rotation angles, $L_s$ and $L_w$ are the regression loss for grasp scores and widths. $\lambda_1$, $\lambda_2$ and $\lambda_3$ are predefined weights for the loss.

%%%%%%%%%%%%%%%%%%%%%%%%%%%%%%%%%%%%%%%%%%%%%%%%%%%%%%%%%%%%%%%%%%%%%%%%%%%%%%%%
\section{EXPERIMENT}

\begin{figure*}[t]
    \centering
    \includegraphics[width=\linewidth]{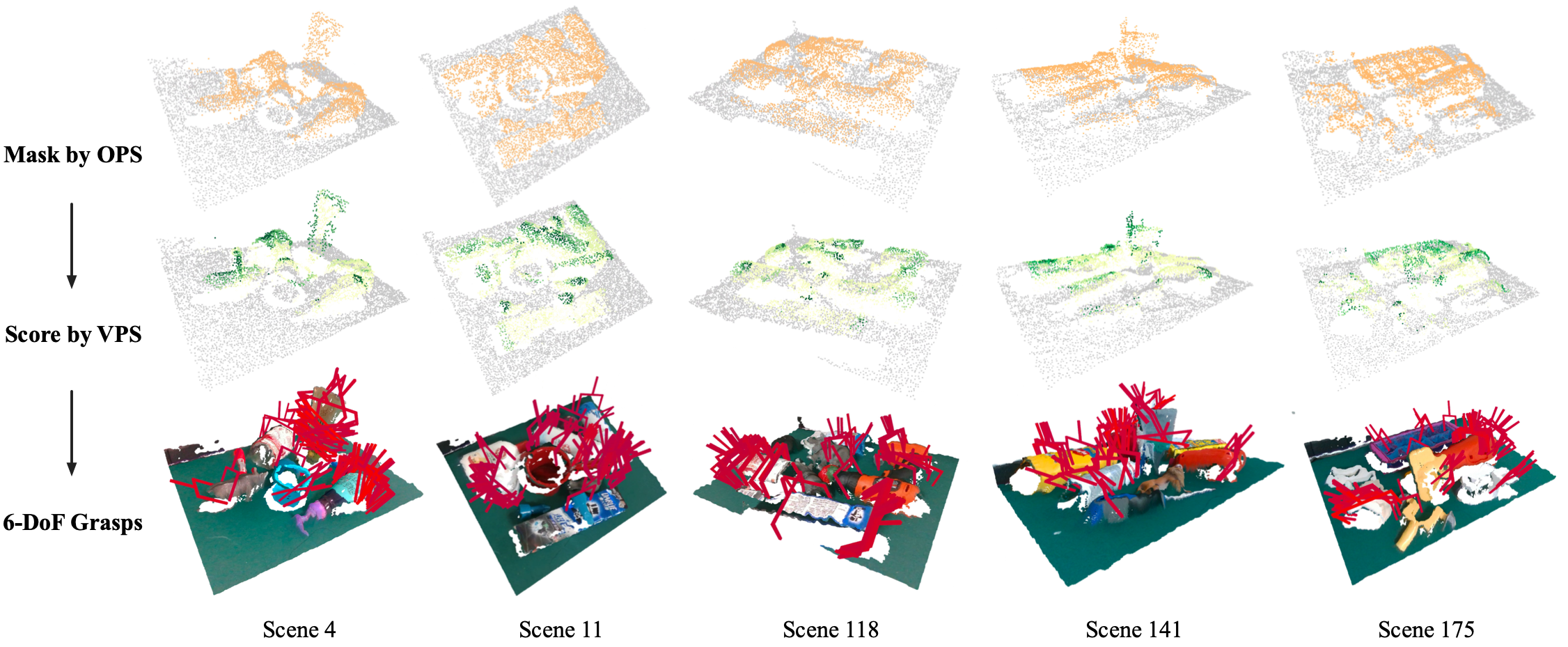}
    \caption{Visualization of grasps detected by GraNet. The top row shows the binary classification results of OPS, where object surface points are marked in orange. The middle row demonstrates the prediction of VPS in green, where darker green means a higher grasp value for that location. The bottom row visualizes the grasps conducted in the corresponding scenario, where the grasp points are obtained by OPS and VPS.}
    \label{fig:result}
    \vspace{-1em}
\end{figure*}

\subsection{Experiment Setup}

\subsubsection{Dataset and Metrics}

We conduct experiments on GraspNet-1Billion \cite{fang_graspnet-1billion_2020} dataset, a large-scale benchmark containing 97,280 RGB-D images of 190 cluttered scenes. GraspNet-1Billion provides over one billion grasp pose annotations for 88 objects, enabling our network to learn grasp configurations in a larger grasping domain.
% 190 scenes are split into 4 parts: 100 scenes for training, 30 scenes for testing seen objects, 30 scenes for testing similar objects, and 30 scenes for testing novel objects.
The evaluation metric \textit{Precision@}$k$ \cite{fang_graspnet-1billion_2020} measures the precision of top-$k$ ranked grasps in a scene, which describes the percentage of true positive grasps. $\mathrm{AP}_\mu$ is the average \textit{Precision@}$k$ under friction $\mu$. In practice, we set $k=50$. $\mathrm{AP}$ is the average of $\mathrm{AP}_\mu$ where $\mu$ ranges from 0.2 to 1.2, with 0.2 as the interval.

\subsubsection{Implementation Details}

Given a randomly sampled point cloud of size $N=12000$, we first construct a KNN graph where each node is connected to the nearest 32 nodes. In GFE, to cope with textureless targets, the original features of graph nodes are only spatial positions, i.e. $d=3$. We learn local graph structure within hop $p=4$ while embedding feature to dimension $f=64$ in each module. To learn more efficiently, we then sample the graph to $7000$ points by FPS and adopt a single EdgeConv of size $(64\times3, 256)$ to learn graph features. In GPS, we get object mask by $\zeta_{obj}$ with the size of $(256, 128, 64, 16, 2)$ and select $N_{obj}=2048$ surface points by FPS. In VPS, $\mathcal{F}$ and $\zeta_{val}$ have size of $(256, 256)$ and $(256, 64, 32, M)$ in order to divide the scores into $M=10$ levels. $N_{val}=512$ points are finally selected as grasp points. We set $\lambda_1, \lambda_2, \lambda_3 = 0.5, 0.3, 0.2$ for loss function.

Our network is trained on a single NVIDIA RTX 3080 GPU (10GB) with only \textbf{30\% of the training data} in the GraspNet-1Billion dataset. We trained our network for 10 epochs of batch size 2 with Adam optimizer. The initial learning rate is 0.001 and decays to 5e-4 after 8 epochs.

\subsection{Main Results on GraspNet-1Billion}

% Overview: conditions
We compare our network to different methods on GraspNet-1Billion. Results are shown in TABLE. \ref{tab:result}. GraNet outperforms traditional 3-DoF methods and recent 6-DoF networks without introducing texture or semantic information, achieving state-of-the-art performance.
% Meanwhile, GraNet does not introduce new grasp annotations or modify the posterior grasp configuration procedure, which proves that our multi-level graph network and learning-based grasp points selection strategy is effective for grasp pose detection problems.

% Numerical comparison
% Simultaneous: 6.78/11.62/0.89 -> 19/41/6
% TransGrasp: 3.52/10.66/1.07 -> 9/36/8
\subsubsection{Metrics Evaluation}
Our method boosts $\mathrm{AP}$ by \textbf{6.78\,/\,11.62\,/\,0.89} and \textbf{3.52\,/\,10.66\,/\,1.07} compared to most recent works Li \textit{et al.} \cite{li_simultaneous_2021} and TransGrasp \cite{liu_transgrasp_2022} on seen\,/\,similar\,/\,novel splits respectively, while training with a small amount of training data. In particular, our method significantly improves the grasping quality on the unseen split with multiple friction factors (\textit{e.g.} $\mathrm{AP}$\,+\,\textbf{41\%} compared to \cite{li_simultaneous_2021} and $\mathrm{AP}$\,+\,\textbf{36\%} compared to \cite{liu_transgrasp_2022}), indicating that our network efficiently learns the geometric features in the training set and can exhibit high performance grasping on similar but unseen objects. We argue that graph-based scene representation emphasizes edge features between nodes and thus has a more powerful geometric feature learning capability. At the same time, the learning-based grasp points selection method further binds the graph structure and graph network to the grasping problem, forming a multi-graph hierarchy.

\subsubsection{Ablation Studies}
To examine the effectiveness of each subnetwork, we perform an ablation experiment by replacing the GPS network with FPS sampling. As can be seen from TABLE. \ref{tab:result}, our learning-based strategy for selecting grasp points significantly improves the precision of grasping (by more than 10 $\mathrm{AP}$), demonstrating the functionality of GPS in locating high-quality grasps. 
In addition, the grasp effect already surpasses many baseline methods when using only the GFE network. This indicates the superiority of graph structure and graph network in dealing with the grasp pose generation problem.

\subsubsection{Visualization Analysis}
We select some example scenes from different data split to illustrate the performance of GraNet. It can be seen from Fig. \ref{fig:result} that OPS can precisely identify foreground targets while VPS can select high-value locations on the target surface to perform grasping. Unlike previous methods, the grasps generated by GraNet are not uniformly distributed but are concentrated in the most favorable regions for grasping  (\textit{e.g.} grasps are always located on objects with a certain height or with a large space around them in Scene 4).

\subsection{Robotic Experiments}
% AMD Ryzen 7 5800X 8-Core Processor
% NVIDIA RTX 3080 GPU
% Ubuntu 20.04

% Hardware-setup
The real-world experiments are conducted on a UR5 robotic arm  
with an attached two-finger gripper. A Photoneo PhoXi 3D scanner is mounted at a fixed place to capture partial scene point clouds.
% Scene
We evaluate GraNet in four cluttered scenes and randomly place 3 to 5 objects to compose each scene. To verify the generalization ability of our object-agnostic grasping method, we conducted experiments using only \textit{novel} objects, which never appeared in the training set.
% Grasps
The robotic arm performs the best grasping prediction each time until all the objects in the scene are grasped, and the max number of grasping attempts is set to 7.
% Figure illustration
The robotic platform and objects are shown in Fig. 5.

\begin{figure}[h]
    \centering
    \includegraphics[width=\linewidth]{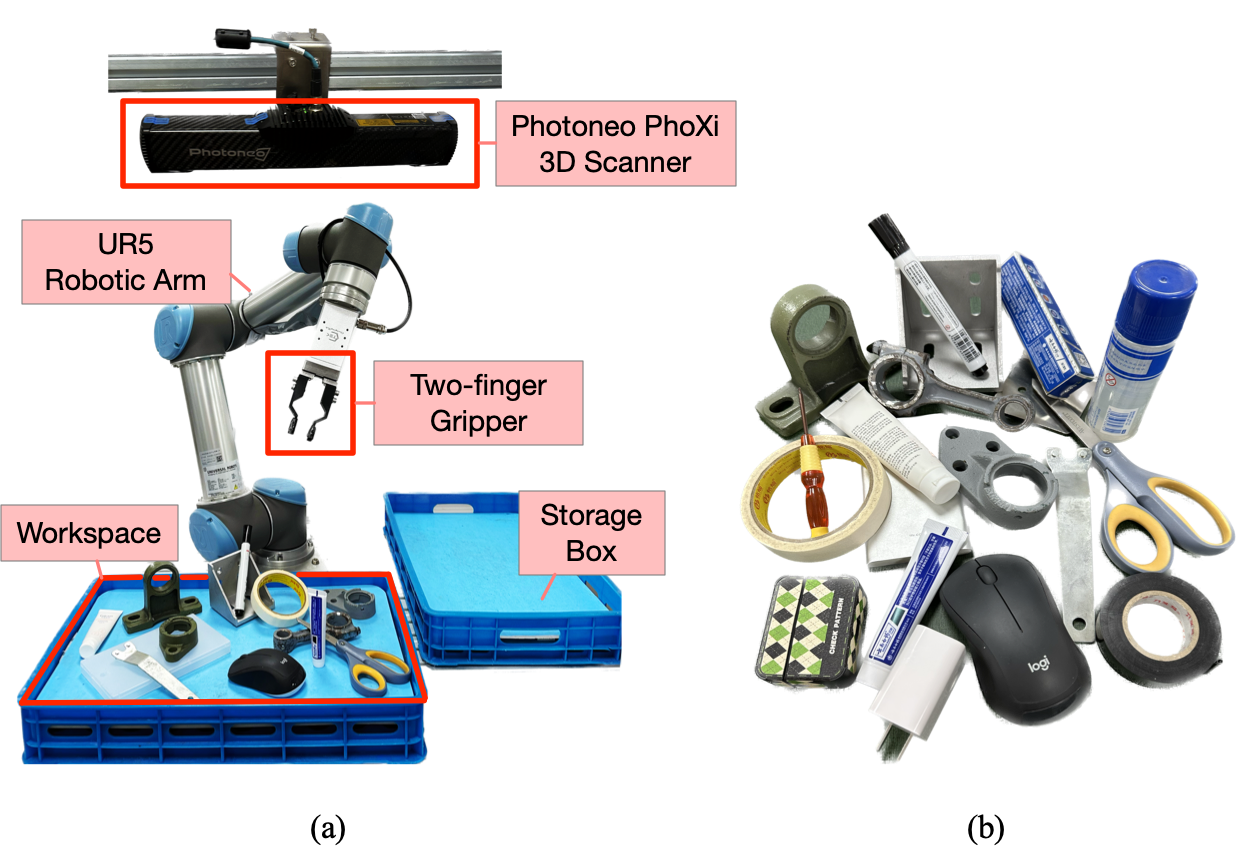}
    \vspace{-1.5em}
    \caption{Robotic experiment platform. (a) Experiment setup. (b) Objects used in our experiment.}
    \label{fig:my_label}
\end{figure}

We report the success rate (SR) and completion rate (CR) in TABLE. 2. Even though we use a different depth sensor than the training set, our method still performs well when there are differences in data distribution. The real-world experiments demonstrate the effectiveness and applicability of our grasping framework in scattered scenarios.

\renewcommand\arraystretch{1.1}
\begin{table}[h]
    \centering
    \caption{Results of cluttered scenes experiment on novel objects}
    \label{tab:robotic}
    \begin{tabular}{c|c|c|c|c}
    \hline
       Scene & \#\,Objects & \#\,Attempts & SR & CR \\
       \hline
       Scene1 & 4 & 5 & 80\% & 100\% \\
       Scene2 & 4 & 4 & 100\% & 100 \\
       Scene3 & 5 & 7 & 71.4\% & 80\% \\
       Scene4 & 3 & 3 & 100\% & 100\% \\
       \hline
       Average & 4 & 4.75 & 84.2\% & 93.75\% \\
       \hline
    \end{tabular}
    \vspace{-1em}
\end{table}

%%%%%%%%%%%%%%%%%%%%%%%%%%%%%%%%%%%%%%%%%%%%%%%%%%%%%%%%%%%%%%%%%%%%%%%%%%%%%%%%
\section{CONCLUSIONS}

In this work, we present an end-to-end 6-DoF grasp pose generation framework to model the complex mapping $\mathcal{M}: \mathcal{P}\rightarrow\mathcal{G}$.
Inputting a point cloud scene, the proposed GraNet captures geometric cues through an attention-based graph network and selects optimal grasping locations by a learning approach.
We deeply integrate the grasping task into feature learning networks and thus form a more targeted understanding of the scene.
Our method outperforms existing works on the GraspNet-1Billion dataset while using less training data. In the experiment of grasping similar objects, GraNet improves the performance by 41\% compared with the current advanced method. We believe that the construction of multi-level graphs enables GraNet to obtain powerful geometric reasoning capabilities. The promising results in robot experiments also indicate the effectiveness and applicability of our work.


%%%%%%%%%%%%%%%%%%%%%%%%%%%%%%%%%%%%%%%%%%%%%%%%%%%%%%%%%%%%%%%%%%%%%%%%%%%%%%%%
\section*{ACKNOWLEDGMENT}
% 自然科学基金面上项目
This work is supported in part by the National Natural Science Foundation of China under Grant No. 52275500.

%%%%%%%%%%%%%%%%%%%%%%%%%%%%%%%%%%%%%%%%%%%%%%%%%%%%%%%%%%%%%%%%%%%%%%%%%%%%%%%%
% 参考文献（IEEETran）
\bibliographystyle{IEEEtran}
\bibliography{references}

\end{document}